\documentclass{article}
\usepackage{spconf,amsmath,graphicx}
\usepackage[utf8]{inputenc}
\usepackage{hyperref}
\usepackage{xcolor}
\definecolor{Green}{rgb}{0.0, 0.42, 0.24} % color

\title{Liver Lesion Segmentation with slice-wise 2D Tiramisu and Tversky loss function}
\name{Karsten Roth \qquad Tomasz Konopczyński \qquad Jürgen Hesser }
\address{Experimental Radiation Oncology, Department of Radiation Oncology,\\
University Medical Center Mannheim, Heidelberg University, Germany}
\begin{document}
%\ninept
%
\maketitle
\begin{abstract}
% The goal of the LiTS (Liver Tumor Segmentation Challenge) competition is to compare methods for automatic or semi-automatic segmentation of liver lesions in CT scans.
% %taken from a diseased population.
% At the present, lesion segmentation is still performed manually (or semi-automatically) by medical experts.
% %The segmentation of liver lesions is a biomedical problem which in the majority of cases requires a manual or semi-automatic procedure led by respectively educated medical personnel.

% The results provided here are in parts based on the conclusions drawn from the participation at the preceding competition version aimed for ISBI 2017 where the viability of a standard slice-wise \textit{U-Net} architecture was investigated. For MICCAI 17, a pipeline using a \textit{Tiramisu} network and Tversky-coefficient based loss was contributed that achieves very good shape extractions with high detection sensitivity.
% Again, our method is fully automatic. We break down the segmentation into two steps: we first segment the liver (ignoring lesions), then perform a lesion segmentation and liver-mask the final result. This provides a good network-only pipeline that can be improved on by various post-processing methods.

At present, lesion segmentation is still performed manually (or semi-automatically) by medical experts. To facilitate this process, we contribute a fully-automatic lesion segmentation pipeline.
This work proposes a method as a part of the LiTS (Liver Tumor Segmentation Challenge) competition for ISBI 17 and MICCAI 17 comparing methods for automatic segmentation of liver lesions in CT scans. 

By utilizing cascaded, densely connected 2D U-Nets and a Tversky-coefficient based loss function, our framework achieves very good shape extractions with high detection sensitivity, with competitive scores at time of publication.
In addition, adjusting hyperparameters in our Tversky-loss allows to tune the network towards higher sensitivity or robustness.
The implementation of our method is available  online\footnote{\url{https://github.com/Confusezius/unet-lits-2d-pipeline}}.
%The latest implementation\footnote{\url{https://github.com/Confusezius/unet-lits-2d-pipeline}} can be found under with better setup and improved results.

\end{abstract}
\section{Introduction}
\label{sec:intro}

As neural networks have proven to achieve human-like performance in biomedical image classification \cite{DERM}, and nearly human-like performance for segmentation tasks \cite{SEG}, it stands to reason to construct a segmentation pipeline that applies these methods for automatic lesion detection.

We evaluate what results can be achieved using the deep fully convolutional neural network architectures \textit{Tiramisu} \cite{tiramisu} and the \textit{U-Net} \cite{UNET} with standard data preprocessing and data augmentation and without any sophisticated postprocessing. 

\textit{U-Net}-based architectures themselves are widely used for biomedical image segmentation, and have already proven its robustness and strength in many challenges ($e.g.$ segmentation of neuronal structures \cite{NEURO}). The \textit{Tiramisu} \cite{tiramisu} enhances this structure on the basis of densely connected networks created by \cite{densenet} to achieve better information flow and higher effective depth. \\
The complete pipeline resembles the cascaded approach proposed in \cite{LITS} where the authors suggested a \textit{U-Net} cascade (with an additional subsequent postprocessing step based on conditional random fields). In our case for liver segmentation, a standard \textit{U-Net} was trained to reduce overall training time, whereas for lesion segmentation, the aforementioned \textit{Tiramisu} was trained from scratch (without transfer learning) with a Tversky loss function to improve on a simple Dice-based loss function.

\section{Dataset and Preprocessing}
\label{sec:prepro}

\begin{table*}[h]
\centering
\begin{tabular}{c|c|c|c|c|c|c|c}

\textbf{Architecture} & \textbf{Dice (A)} & \textbf{Dice (G)} & \textbf{VOE} & \textbf{RVD} & \textbf{ASSD} & \textbf{MSSD} & \textbf{RMSD}\\
\hline
\textit{U-Net LiS (MICCAI 17)}& 0.95 & 0.94 & 0.10 & -0.05 & 1.89 & 32.71 & 4.20\\
\hline
\textit{U-Net + U-Net LeS (ISBI 17)}& 0.42 & - & 0.69 & 103.74 & 32.54 & - & -\\
\textit{U-Net + Tiramisu (DL) LeS (MICCAI 17)} & 0.45 & 0.44 & 0.33 & -0.16 & 1.28 & 8.85 & 2.07\\
\textit{U-Net + Tiramisu (TL) LeS (MICCAI 17)} & 0.57 (14)& 0.66 (16)& 0.34 (1)& 0.02 (4)& 0.95 (1)& 6.81 (6)& 1.60 (3)\\
\hline
\end{tabular}
\caption{Test results for various pipeline setups. Comparison metrics are Dice (Average/Global), Volume-Overlap Error (VOE), Relative Volume Differene (RVD), Average and Maximum Symmetric Surface Distance (A/MSSD) and Root-Mean-Square-Deviation (RMSD). DL denotes the network being trained with dice loss, and TL with Tversky loss respectively. LiS stands for liver segmentation and LeS for lesion segmentation. Parenthesis denote the standings in the final ranking for each metric. MICCAI 17/ISBI 17 mark the competition the segmentations were submitted to. }
\label{tab:res}
\end{table*}

The dataset consists of 200 CT-scans
of the abdominal and upper body region such that the liver is fully included as part of the Liver Tumor Segmentation Challenge \cite{litsnew}.
The data was provided by different clinics from all over the world
and is stored in NIfTI dataformat.
For working purposes, the set was divided by the LITS challenge organisers
into 130 scans for training (containing liver and lesion masks)
and 70 for evaluating the method (without masks).
From the 130 training scans, we selected 25 scans for validation purposes during
training and 105 for actual training.
% What is the "standard" preprocessing? Do you mean this is what you have
% described in more detail below? I wouldn't write "standard". Just say
% As preprocessing, we perform the following steps: 1, 2, 3...

Preprocessing on the provided data included value-clipping the volume values (on the Hounsfield scale) to an interval of $[-100,400]$ in order to remove air- and bonelike structures for a more uniform background. The voxel values were then normalized by substracting the training set mean and standardized by dividing through the training set standard deviation. In a final step, the affine matrix was extracted from the NIfTI header to rotate each volume to the same position to make learning easier.

\section{Method}
\label{sec:method}

As mentioned, the lesion segmentation pipeline includes two different nets:
a \textit{U-Net} to segment the liver from a given volume slice,
and the \textit{Tiramisu} to segment out lesions.
By masking the lesion segmentation with the liver mask, we effectively reduce the number of false positives in regions outside the liver.
Both architectures will be presented shortly.
We implemented our networks by use of the deep learning library Keras \cite{KERAS} on the basis of tensorflow \cite{tensorflow}.

\subsection{U-Net architecture}
Our \textit{U-Net} was implemented with two convolutional layers with padding plus a ReLU-activation layer
before max-pooling for the downward movement of the U-branch.
These blocks were repeated for a total of four times.
%We started with a filter number of 32 and constant 3$\times$3 filter size.
%The filter number was doubled after every convolution-max-pooling-block.
We started with 32 filters and a constant 3$\times$3 filter size.
The number of filters was doubled after every convolution-max-pooling-block.
Dropout was performed after the ReLU-activation layer with a rate of 0.2.
The upsampling branch which performs a transposed convolution to reconstruct
the segmentation image was implemented in the same fashion. This totals in roughly $7.7\cdot 10^6$ network parameters.

\subsection{Tiramisu architecture}
The \textit{Tiramisu} was implemented as described in the original paper with 4 denseblocks comprising 4,5,6 and 7 layers respectively in the feature extraction branch and a final bottleneck denseblock with 8 densely connected layers. Each denseblock was created with a growth rate of 12, followed by a maxpooling layer while initializing the network with 32 starting filters. Dropout with probability 0.2 was performed after each denseblock, and the extraction structure was mirrored for the upsampling step. ReLU activation was used and L2-regularization with $\lambda=10^{-5}$ was added to the convolutional layers. In addition, a constant $3\times 3$ filter size was implemented. All in all, this results in approx. $1.8\cdot 10^6$ parameters, i.e. only a fourth of the standard \textit{U-Net} setup that was used for liver segmentation/ISBI 17. Standard Batch-Normalization was applied before each convolutional layer.

\subsection{Liver segmentation training}
Training was performed over 50 epochs with a batch size of 5 on a GTX 1070 GPU using an Adam optimizer \cite{ADAM}, He uniform initialization \cite{heinit}, standard binary crossentropy loss and a learning rate of $10^{-5}$ which was halved every 15 epochs. The final weights were selected to provide the best validation score. In total, training on 105 randomly selected volumes (including validating on 25) took approximately 80 hours. Note that no data augmentation was performed.

\subsection{Lesion segmentation training}
The \textit{Tiramisu} lesion segmentation network was trained for 35 epochs using the Adam optimization method and a tversky loss function \cite{tversky} based on the Tversky coefficient (through simple negation).

For the loss, the coefficient was negated with false positive penalty  $\beta = 0.7$ and false negative penalty $\alpha=0.3$. We penalize false positive predictions higher because training the \textit{Tiramisu} without transfer learning and simple weighted binary cross entropy or dice loss showed the network being prone to a high false positive rate. An initial learning rate of $3\cdot 10^{-6}$ was chosen, which was halved every 10 epochs. The complete training using a batch-size of 5 required nearly 38 hours, with the usage of very basic data augmentation (minor rotation and translation as well as zooming).
During training, the network learned from slices along the coronal axis of which $224\times 224$ crops were taken in and around the liver. In different runs (see tab.\ref{tab:res}), a simple Dice loss was also used for training as reference to the Tversky loss under otherwise same conditions.

\section{Experiment and results}
\label{sec:method}

For the final submission, after each liver segmentation, only the largest connected component was chosen as the final liver mask. In addition to the test results for the introduced pipeline, tab.\ref{tab:res} also shows the results for the \textit{U-Net}-only approach for ISBI 17, and a pipeline with a \textit{Tiramisu} lesion segmentation network that was trained with a simple dice loss function. Each method was applied on the 70 test volumes and shared the same liver segmentation network.
Total inference time including both liver and lesion segmentation took roughly 7s per volume on average, totaling in 8 hour of testing time on a GTX 1070 GPU.
For a measurement of tumorburden, the pipeline scores 0.03 on RSME and 0.18 on Max Error.

\begin{figure*}[htb]
\includegraphics[width=1\linewidth]{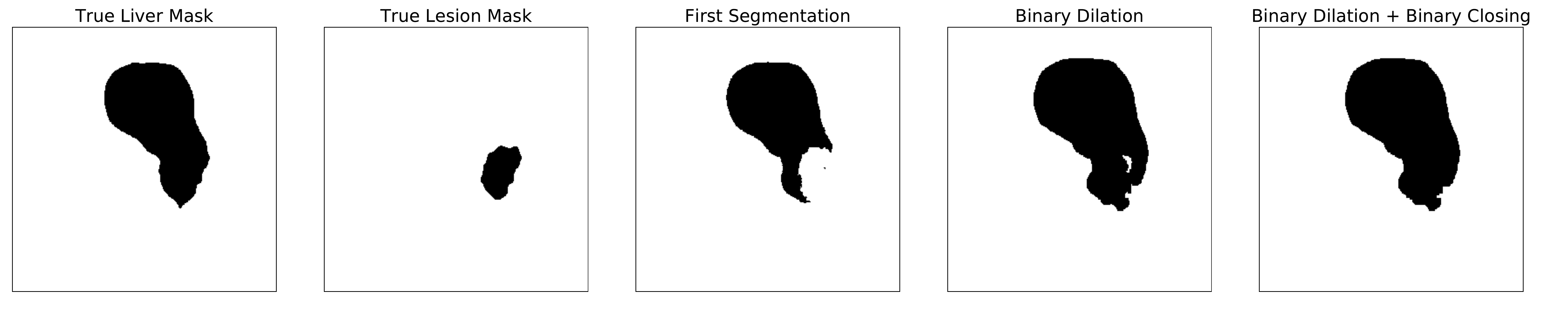}
\caption{From left to right: Ground truth liver mask, Ground truth lesion mask, \textit{U-Net}-segmented liver mask, mask after binary dilation with structure element of size $7\times 7\times 7$ and after binary dilation and closing with same-sized structure elements.}
\label{fig:bad_seg}
\end{figure*}

% Add example image and slice

\begin{figure}[t]
\begin{minipage}[b]{0.48\linewidth}
  \centering
  \centerline{\includegraphics[width=3.8cm]{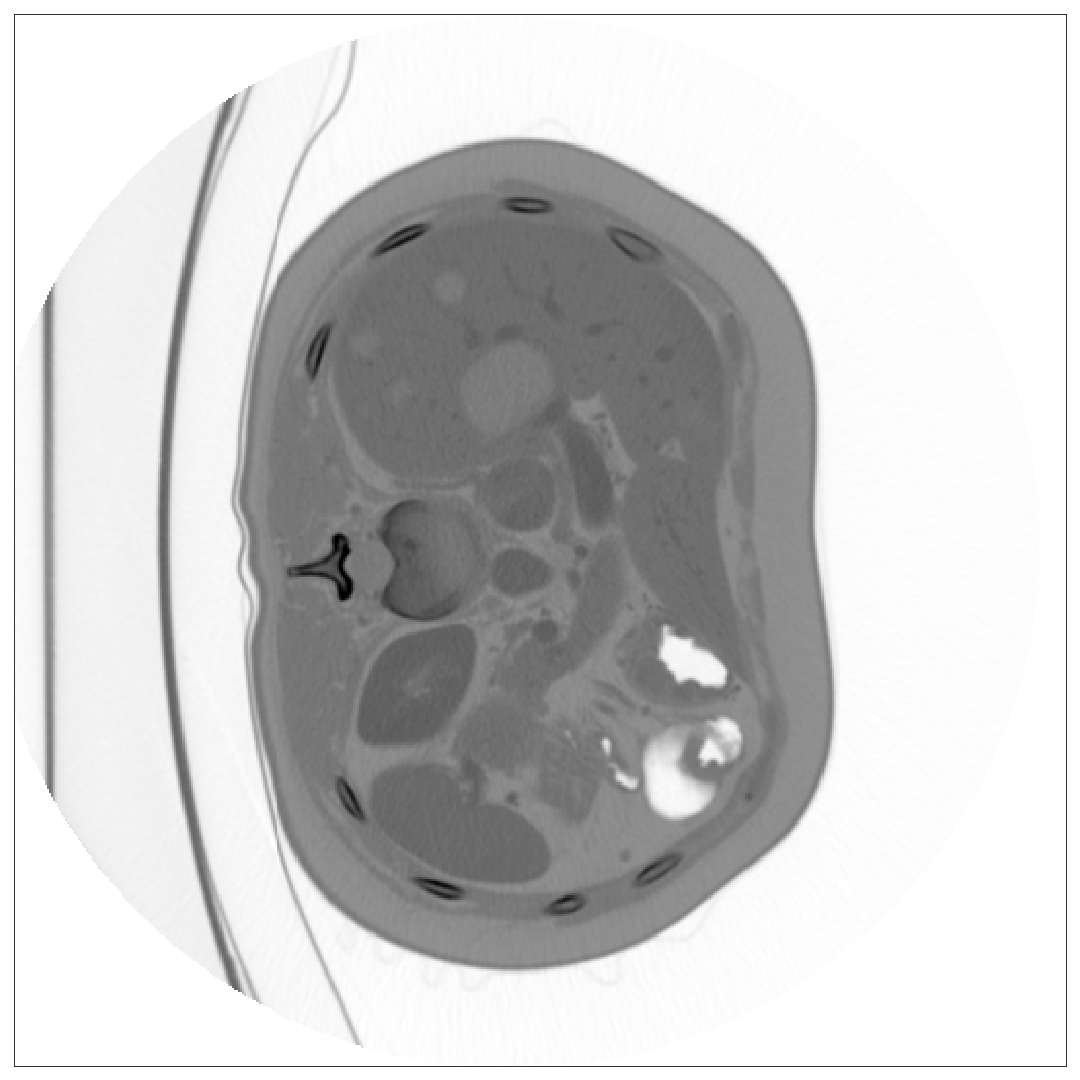}}
%  \vspace{1.5cm}
  \centerline{(a) Example slice}\medskip
\end{minipage}
\hfill
\begin{minipage}[b]{0.48\linewidth}
  \centering
  \centerline{\includegraphics[width=4.0cm]{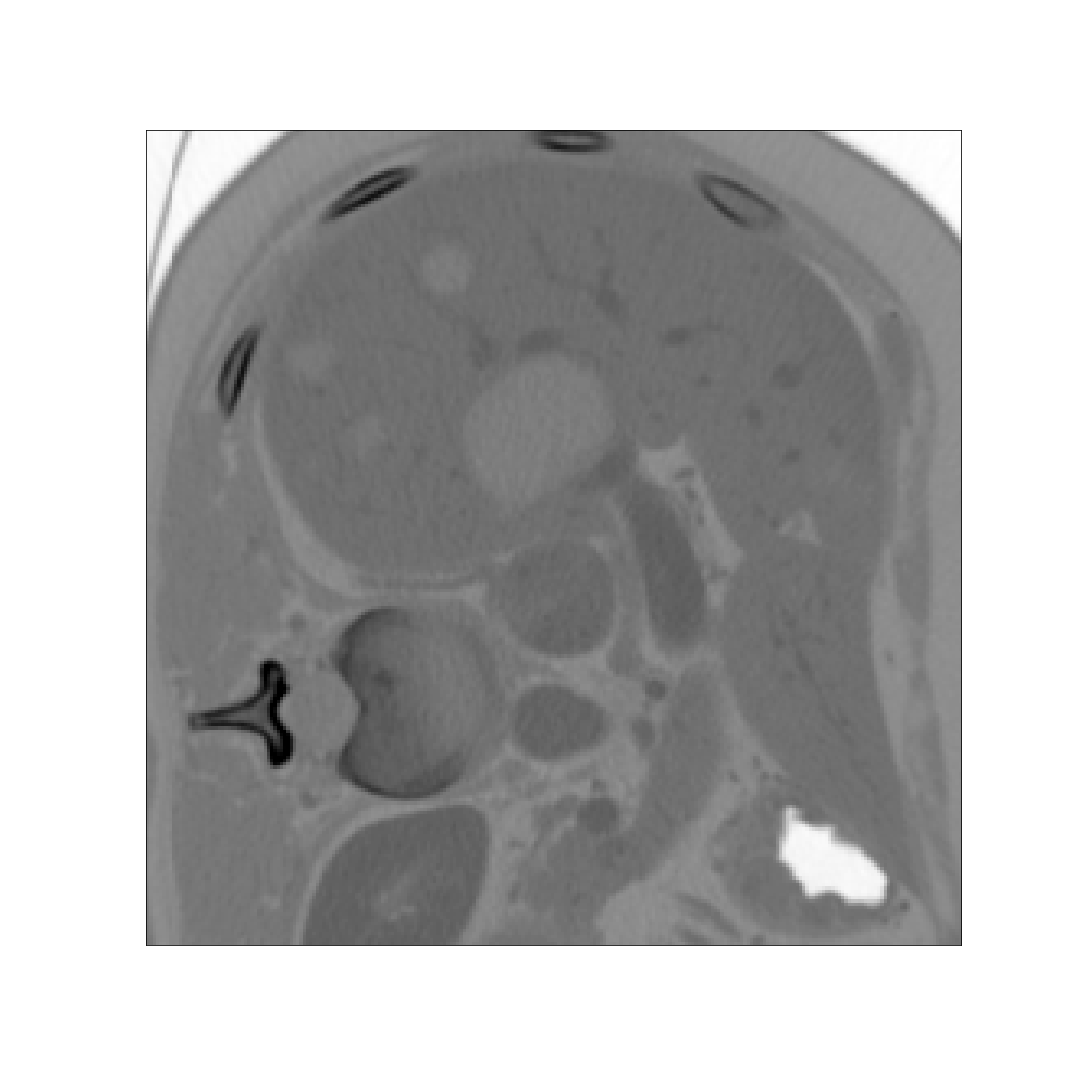}}
%  \vspace{1.5cm}
  \centerline{(b) Zoomed Liver}\medskip
\end{minipage}
\hfill
\begin{minipage}[b]{.48\linewidth}
  \centering
  \centerline{\includegraphics[width=4.0cm]{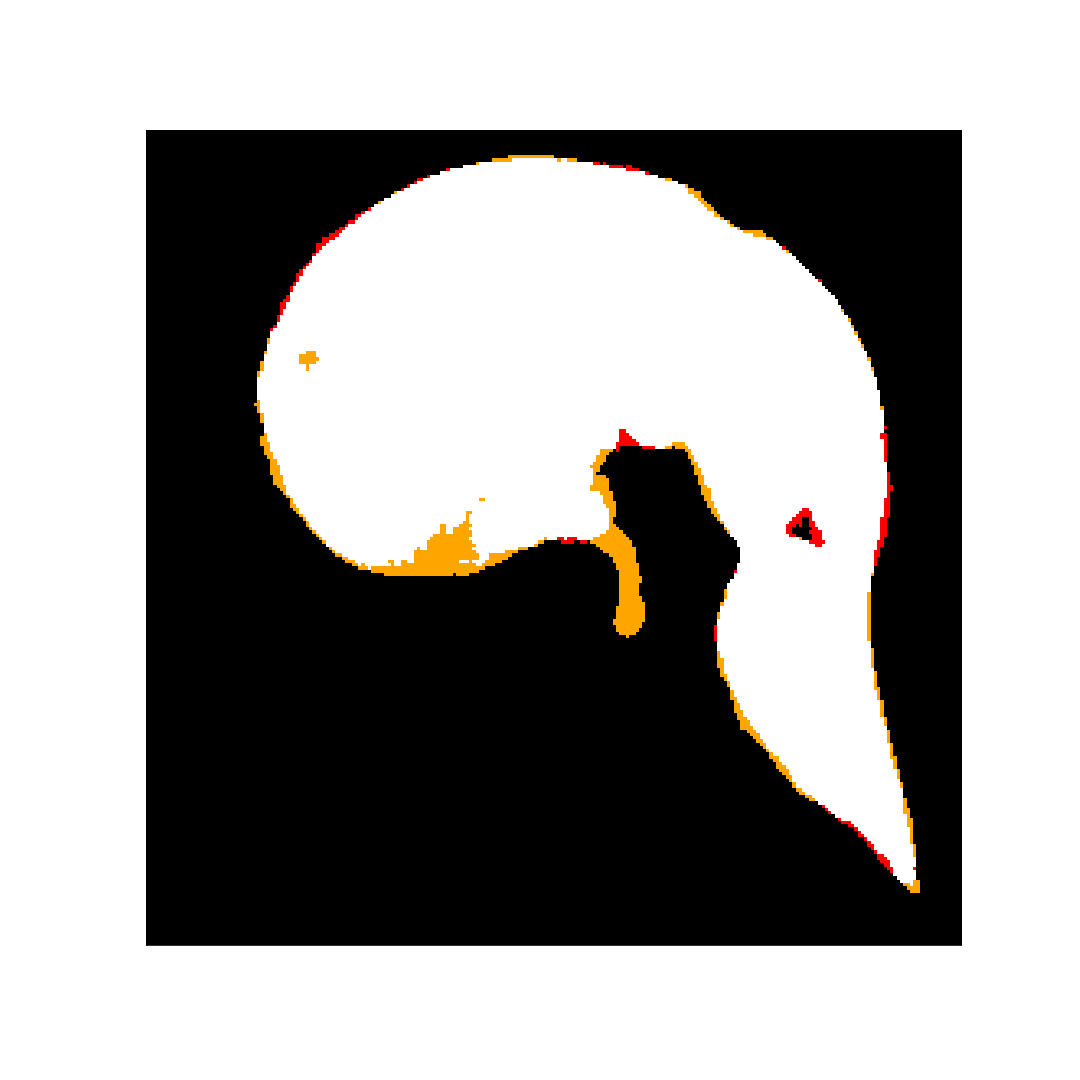}}
%  \vspace{1.5cm}
  \centerline{(c) Liver Segmentation}\medskip
\end{minipage}
\hfill
\begin{minipage}[b]{0.48\linewidth}
  \centering
  \centerline{\includegraphics[width=4.0cm]{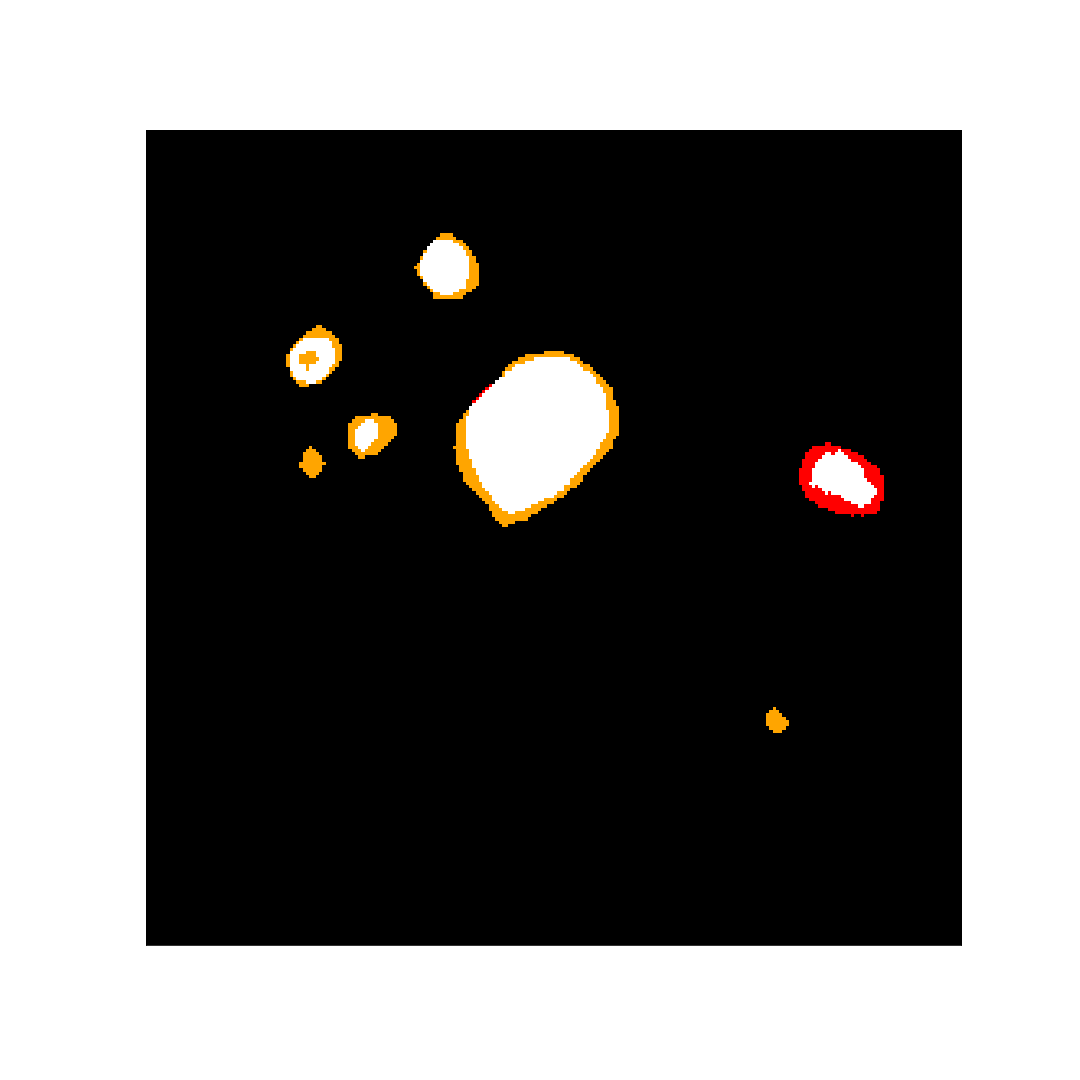}}
%  \vspace{1.5cm}
  \centerline{(d) Liver lesion segmentation}\medskip
\end{minipage}
\caption{Segmentation of liver and liver lesion from an example slice (a) from the validation data set (volume 46 from the LiTS challenge). The resulting segmentation images (c) and (d) are color coded, where the white color means True Positive, red color False Positive and orange color False Negative.}
\label{fig:segmentation}
\end{figure}

\section{Discussion}
\label{sec:StrWeak}

The major strength of this approach is its simplicity, as we use a network-only based segmentation approach to achieve good segmentation results (see tab.\ref{tab:res}), especially for the \textit{Tiramisu}-based pipeline when compared to a plain \textit{U-Net} setup.

Moreover, the proposed pipeline using \textit{Tiramisu} and Tversky-loss achieves best scores regarding Volume-Overlap Error (VOE) and Average Symmetric Surface Distance (ASSD) as well as scoring high in Relative Volume Difference (RVD) and Maximum Symmetric Surface Distance (MSSD). This insinuates that the Tiramisu performs very well when it comes to actual shape extraction when lesions are detected correctly.

However, there are several issues solving which could improve the segmentation capabilities:\\
\textbf{(1)} As can be seen from fig.\ref{fig:segmentation} and tab.\ref{tab:res}, our liver segmentation performs reasonably well, but misses a majority of very big and/or lesion at the liver boundary (see fig.\ref{fig:bad_seg}) which then, although maybe detected through the lesion segmentation network, get masked away. Increasing the training effort for our liver segmentation network would therefore allow us to better our current lesion (and liver) segmentation results. We tried to counter this effect by isotropically dilating the liver mask (fig.\ref{fig:bad_seg}), but this only worsened the final results as the number of introduced false positives was higher then the amount of removed false negatives.\\
\textbf{(2)} Even with our current implementation of the Tversky loss function, the major error source are false positive prediction. Higher penalties, a network ensemble and regularization through, for example, higher dropout or more sophisticated data augmentation would ideally increase the area under the ROC curve and provide better overall scores.

In addition, training a liver segmentation network on the basis of a \textit{Tiramisu} and using that as weight initialization could potentially help the network to converge faster and to better results.

On a final note, \cite{LITS} and \cite{unet_rf} showed that good postprocessing (for example with 3D conditional random fields and random forests with additional handmade features) could improve the result. We feel this is an avenue worth exploring on this data.
Also, segmentation was only performed on a 2D slice-by-slice basis. \cite{2p5Dnet} showed exceptional results through the application of a semi-3D approach by including additional upper and/or lower slices in the segmentation process of the neural network to incorporate more spatial information. Our pipeline could be easily extended to make use of this approach to hopefully improve the results further.

\newpage

\bibliography{asd} 
\label{sec:ref}
\bibliographystyle{IEEEbib}

\end{document}